\documentclass[11pt]{article}
\usepackage[all]{xy}
\usepackage{amssymb}
\usepackage{multirow}
\usepackage{amsmath}
\usepackage{amsfonts}
\usepackage{diagbox}
 \usepackage{pifont}
\usepackage{amsmath,amsfonts,amssymb,theorem,graphicx,listings,txfonts}
\usepackage{graphics}
 \usepackage{caption2}
\usepackage{flafter}
 \usepackage{indentfirst}
\usepackage{epsfig}
  \usepackage{mathrsfs}
  \usepackage{float}
\usepackage{subfigure}
 \usepackage{cite}
\newtheorem{theorem}{Theorem}[section]

\newtheorem{definition}[theorem]{Definition}

\theoremstyle{example}
\newtheorem{example}[theorem]{Example}
\theoremstyle{programme}

\theoremstyle{property}

\theoremstyle{problem}

\renewcommand{\arraystretch}{1}

\title{Three-Way Decisions-Based Conflict Analysis Models}

\author
{Guangming Lang$^{1,2}$ 
\thanks{Corresponding author.\quad
\newline\mbox{}\hspace{0.55cm}
E-mail address: langguangming1984@126.com(G.M.Lang). }\hspace{1cm}\\
\small {$^{1}$ School of Mathematics and Statistics, Changsha University of Science and Technology}\\
\small {Changsha, Hunan 410114, P.R. China}\\
\small {$^{2}$Department of Computer Science, University of Regina}\\
\small {Regina, Saskatchewan, S4S 0A2, Canada}}
\date{}

\begin{document}
\maketitle \baselineskip=17pt
\begin{center}
\begin{quote}
{{\bf Abstract.}
Three-way decision theory, which trisects the universe with less risks or costs,  is considered as a powerful mathematical tool for handling uncertainty in incomplete and imprecise information tables, and provides an effective tool for conflict analysis decision making in real-time situations. In this paper, we propose the concepts of the agreement, disagreement and neutral subsets of a strategy with two evaluation functions, which establish the three-way decisions-based conflict analysis models(TWDCAMs) for trisecting the universe of agents, and employ a pair of two-way decisions models to interpret the mechanism of the three-way decision rules for an agent. Subsequently, we develop the concepts of the agreement, disagreement and neutral strategies of an agent group with two evaluation functions, which build the TWDCAMs for trisecting the universe of issues, and take a couple of two-way decisions models to explain the mechanism of the three-way decision rules for an issue. Finally, we reconstruct Fan, Qi and Wei's conflict analysis models(FQWCAMs) and Sun, Ma and Zhao's conflict analysis models(SMZCAMs) with two evaluation functions, and interpret FQWCAMs and SMZCAMs with a pair of two-day decisions models, which illustrates that FQWCAMs and SMZCAMs are special cases of TWDCAMs.

\\}
\end{quote}
\end{center}
\renewcommand{\thesection}{\arabic{section}}

\section{Introduction}

Three-way decision theory\cite{Yao1}, which
promotes thinking and problem solving in threes such as three regions, three levels and three stages, is regarded as one of leading theories for handling uncertainty in decision making problems. The intrinsic ideas of three-way decision theory\cite{Chen1,Deng,Fujita,Luo1,Hu1,Ma1,Lang1,Zhao1,Azam1,Yao3,Yang1,Yang2,Yan,Zhang1,Zhang2} have been widely applied to many fields such as medical decision-making and
recommender systems. Especially, it has had great success in building classifiers for classifying test examples into three classes of acceptance, non-commitment and rejection. For example, Chen et al.\cite{Chen1} developed the three-way decision support method for handling the uncertain medical cases and achieved precise classification of Malignant Focal Liver Lesions to support liver cancer diagnosis. Hu and Yao\cite{Hu1} provided structured rough set approximations of sets for three-way decision making in both complete and incomplete information tables.
Luo et al.\cite{Luo1} investigated the update problems of three-way decisions with dynamic variation of scales in incomplete multi-scale information systems. Yan et al.\cite{Yan} employed the difference in the cost of selecting key samples to construct a three-way decision ensemble model for imbalanced data oversampling.
Zhang, Min and Shi\cite{Zhang2} established a regression-based three-way recommender system and minimized the average cost by adjusting the thresholds for different behaviors.

Conflicts, as one of the most essential characteristic of human nature, exist extensively due to the scarcity of resources and cultural diversity in social life, and conflict analysis and resolution, which explore the structure of conflicts for making proper decisions, play an important role in many fields such as business, political and legal disputes and military operations. Especially, Pawlak\cite{Pawlak1} initially provided rough sets-based conflict analysis models which trisect the universe of agents with respect to an issue set. Nowadays, researchers\cite{Ali1,Deja,Deja1,Deja2,Jiang,Jabbour,Liu,Lang,Pawlak2,Pawlak3,Sun1,
Sun2,Fan,Ganter,Liu1,Yu1,Ramanna1,Silva,Skowron,Skowron1} have established more conflict analysis models for decision making from different views. For example, Fan, Qi and Wei\cite{Fan} employed the including degree to construct a type of TWDCAMs over two universes and divided agents and issues into three disjoint blocks with respect to a strategy and an agent group, respectively. Lang, Miao and Cai\cite{Lang} developed another type of TWDCAMs and studied how to divide agents into three disjoint blocks in dynamic situation tables.
Sun, Ma and Zhao\cite{Sun1} established rough sets-based conflict analysis models over two universes and gave a convenient way to analyze and solve the conflict situation. Furthermore, the existing conflict analysis models provide an effective tool
for decision making problems in conflict situations. Especially, we observe that FQWCAMs divide the universes of agents and issues into three disjoint blocks with the including degree from the view of formal concepts; we also find that SMZCAMs employ the set inclusion and intersection to trisect the universes of agents and issues from the view of rough sets, and there are some similarities and differences between FQWCAMs and SMZCAMs, but the existing results have not illustrated the intrinsic ideas of these conflict analysis models. Actually, there are three values for the opinions of agents on issues in situation tables, which is consistent with the intrinsic idea of three-way decisions, and it is very important to study FQWCAMs and SMZCAMs based on three-way decision theory.

The purpose is to study further TWDCAMs. Firstly, we propose a type of TWDCAMs for trisecting the universe of agents based on two evaluation functions. Concretely, we provide the concepts of
an acceptance evaluation and a rejection evaluation of a strategy by an agent. We propose the concepts of the agreement, disagreement and neutral subsets of a strategy with the acceptance and rejection evaluations, which are considered as the intrinsic idea of the TWDCAMs for trisecting the universe of agents. We employ a pair of two-way decisions models to interpret TWDCAMs, FQWCAMs and SMZCAMs for trisecting agents and find that FQWCAMs and SMZCAMs are special cases of this type of TWDCAMs.
Secondly, we develop another type of TWDCAMs for dividing the universe of issues into three disjoint blocks. Concretely, we propose the concepts of
an acceptance evaluation and a rejection evaluation of an agent group by an issue. We introduce the concepts of the agreement, disagreement and neutral strategies of an agent group with the acceptance and rejection evaluations, which are considered as the intrinsic idea of TWDCAMs for trisecting the universe of issues. We also employ a pair of two-way decisions models to interpret TWDCAMs, FQWCAMs and SMZCAMs for trisecting issues and find that this type of TWDCAMs is a generalization of FQWCAMs and SMZCAMs.

The rest of this paper is organized as follows: Section 2 briefly reviews the concepts of rough sets and three-way decisions. In Section 3, we provide TWDCAMs for dividing the universe of agents. Section 4 develops TWDCAMs for trisecting the universe of issues. In Section 5, we investigate the relationship among TWDCAMs, FQWCAMs and SMZCAMs. We conclude the paper in Section 6.

\section{Preliminaries}

In this section, we briefly review the concepts of rough sets over two universes, three-way decisions, and situation tables.

Suppose $U$ and $V$ are two universes, and $R$ be a binary relation from $U$ to $V$. If there exist $t\in V$ and $s\in U$ such that $(u,t),(s,v)\in R$ for any $u\in U$ and $v\in V$, then $R$ is called a compatibility relation. For any $x\in U, v\in V$, the compatibility relation between $x$ and $v$ is a set-valued mapping $R: U\longrightarrow 2^{V}$. That is, $R(x)=\{v\in V\mid (x,v)\in R, \forall x\in U\}$, where $R(x)$ denotes all elements in $V$ related with $x$.

\begin{definition}\cite{Sun1}
Let $U$ and $V$ be two universes, and $R$ a compatibility relation from $U$ to $V$. Then the lower and upper approximations $\underline{apr}_{R}(X)$ and $\overline{apr}_{R}(X)$ of $X\subseteq V$ are defined as follows:
\begin{eqnarray*}
\underline{apr}_{R}(X)=\{x\in U\mid R(x)\subseteq X\},\hspace{0.25cm} \overline{apr}_{R}(X)=\{x\in U\mid R(x)\cap X\neq \emptyset\}.
\end{eqnarray*}
\end{definition}

The set $\underline{apr}_{R}(X)$ consists of elements of $U$ which are only compatible with those elements in $X$, and the set $\overline{apr}_{R}(X)$ consists of elements of $U$ which are compatible with at least one element in $X$. That is, the sets $\underline{apr}_{R}(X)$ and $\overline{apr}_{R}(X)$, which build the bridge between the universes $U$ and $V$, are interpreted as the pessimistic description and optimistic description of $X$, respectively.

\begin{definition}\cite{Yao1}
Let $U$ be the universe of objects, $V$ the universe of attributes, $P(X|[x]_{B})=\frac{|[x]_{B}\cap
X|}{|[x]_{B}|}$ for $X\subseteq U$ and $B\subseteq V$, where $[x]_{B}=\{y\in U\mid \forall c\in B, c(x)=c(y)\}$, $c(x)$ and $c(y)$ are attribute values of objects $x$ and $y$ on the attribute $c$, and $0\leq \beta
<\alpha\leq 1$. Then the positive, boundary and
negative regions $POS_{(\alpha,\beta)}(X)$, $BND_{(\alpha,\beta)}(X)$ and $NEG_{(\alpha,\beta)}(X)$ of $X$ are defined as follows:
\begin{eqnarray*}
POS_{(\alpha,\beta)}(X)&=&\{x\in U\mid P(X|[x])\geq \alpha\};\\
BND_{(\alpha,\beta)}(X)&=&\{x\in U\mid \beta< P(X|[x])<\alpha\};\\
NEG_{(\alpha,\beta)}(X)&=&\{x\in U\mid P(X|[x])\leq \beta\}.
\end{eqnarray*}
\end{definition}

By Definition 2.2, we divide all objects into the positive, boundary and
negative regions. Concretely, for $x\in U$, if $P(X|[x])\geq \alpha$, then $x\in POS_{(\alpha,\beta)}(X)$; if $\beta< P(X|[x])<\alpha$, then $x\in BND_{(\alpha,\beta)}(X)$; if $P(X|[x])\leq \beta$, then $x\in NEG_{(\alpha,\beta)}(X)$. Furthermore, we derive different rules with the positive, boundary and
negative regions. Concretely, the positive region generates positive rules to make a decision of acceptance; the negative region generates negative rules to make a decision of rejection; the third choice, generated from the boundary regions, makes a decision of non-commitment.

\begin{definition}\cite{Pawlak1}
Let $U$ be the universe of agents, $V$ the universe of issues, the attitude of the agent $x\in U$ to an issue $c\in V$ is interpreted as a function $c: U\longrightarrow V_{c}$, where $V_{c}=\{+1,-1,0\}$, $c(x)=+1$ means the agent $x$ agrees with the issue $c$, $c(x)=-1$ represents the agent $x$ opposes the issue $c$, and $c(x)=0$ represents the agent $x$ is neutral towards the issue $c$.
\end{definition}

For simplicity, we refer to $(U,V)$ as a situation table in the following discussion. A situation table, where all agents are measured by using a finite number of issues and all issue values are taken from $\{+1,0,-1\}$, represents the opinions of all agents to some issues. Furthermore, there are many basic research directions for situation tables, and we only study how to trisect the universes of agents and issues based on the three-way decisions theory in this work.

\begin{example}\cite{Pawlak1}
Table 1 depicts the situation for the Middle East, where $U=\{x_{1}, x_{2}, x_{3}, x_{4}, x_{5},x_{6}\}$ and $V=\{c_{1}, c_{2}, c_{3}, c_{4}, c_{5}\}$, where $x_{1}, x_{2}, x_{3}, x_{4}, x_{5}$ and $x_{6}$ are six agents, $c_{1}, c_{2}, c_{3}, c_{4} $ and $c_{5}$ are five issues. For example, $c_{1}(x_{1})=-1$ denotes the agent $x_{1}$ is against the issue $c_{1}$, and $c_{1}(x_{2})=+1$ denotes the agent $x_{2}$ supports the issue $c_{1}$, and $c_{1}(x_{4})=0$ denotes the agent $x_{4}$ is neutral to the issue $c_{1}$.
\end{example}

\begin{table}[H]\renewcommand{\arraystretch}{1.5}
\caption{The Situation Table for the Middle East.}
\tabcolsep0.37in
\begin{tabular}{ |c| c| c| c |c |c|}
\hline
\diagbox{$U$}{$V$} &$c_{1}$ & $c_{2}$& $c_{3}$& $c_{4}$& $c_{5}$\\\hline
$x_{1}$& $-1$ & $+1$ & $+1$ & $+1$ & $+1$ \\\hline
$x_{2}$& $+1$ & $0$ & $-1$ & $-1$ & $-1$ \\\hline
$x_{3}$& $+1$ & $-1$ & $-1$ & $-1$ & $0$ \\\hline
$x_{4}$& $0$ & $-1$ & $+1$ & $0$ & $-1$ \\\hline
$x_{5}$& $+1$ & $-1$ & $-1$ & $-1$ & $-1$ \\\hline
$x_{6}$& $0$ & $+1$ & $-1$ & $0$ & $+1$ \\
\hline
\end{tabular}
\end{table}

{\bf{Remark:}} In Table 1, $x_{1}, x_{2},
x_{3}, x_{4}, x_{5}$ and $x_{6}$ stands for Israel, Egypt, Palestine, Jordan, Syria and Saudi Arabia, respectively; $c_{1}$ means Autonomous Palestinian state on the West Bank and Gaza; $c_{2}$ denotes Israeli military outpost along the Jordan River; $c_{3}$ stands for Israel retains East Jerusalem; $c_{4}$ means Israeli military outposts on the Golan Heights; $c_{5}$ denotes Arab countries grant citizenship to Palestinians who choose to remain within their borders.

\section{TWDCAMs for Trisecting the Universe of Agents}

In this section, we study TWDCAMs for trisecting the universe of agents.

\subsection{TWDCAMs I}

In this section, we provide a type of TWDCAMs for dividing agents into three disjoint blocks.

\begin{definition}
Let $U$ be the universe of agents, and $V$ the universe of issues, $(L_{a},\preceq_{a})$ and $(L_{r},\preceq_{r})$
are two posets. Then a pair of functions $\nu_{a}: U\times P(V)\rightarrow L_{a}$ and $\nu_{r}: U\times P(V)\rightarrow L_{r}$ is called an acceptance evaluation and a rejection evaluation of a strategy by an agent, respectively.
\end{definition}

For an agent $x\in U$ and a strategy $X\subseteq V$, we refer to $\nu_{a}(x,X)$ and $\nu_{r}(x,X)$ as the acceptance and rejection degrees of the strategy $X$ by the agent $x$, respectively; for the agents $x,y\in U$, if $\nu_{a}(x,X)\preceq_{a} \nu_{a}(y,X)$,
then $x$ is less acceptable than $y$ with respect to $X$; for the agents $x,y\in U$, if $\nu_{r}(x,X)\preceq_{r} \nu_{r}(y,X)$, then $x$ is less rejectable than $y$ with respect to $X$.
Furthermore, if the agent $x$ accepts the strategy $X$, then $\nu_{a}(x,X)$ must be in a certain subset $L^{+}_{a}\subseteq L_{a}$, where $L^{+}_{a}$ stands for the acceptance region of $L_{a}$; if the agent $x$ rejects the strategy $X$, then $\nu_{r}(x,X)$ must be in a certain subset $L^{-}_{r}\subseteq L_{r}$, where $L^{-}_{r}$ denotes the rejection region of $L_{r}$. For simplicity, $L^{+}_{a}$ and $L^{+}_{r}$ denote the designated values for acceptance and designed values for rejection, respectively, in the following.

\begin{definition}
Let $U$ be the universe of agents, $V$ the universe of issues, $L^{+}_{a}$ and $L^{+}_{r}$ the designated values for acceptance and designed values for rejection, respectively, $\nu_{a}$ and $\nu_{r}$ the acceptance evaluation and rejection evaluation of an agent to a strategy, respectively, and $X\subseteq V$ a strategy. Then we define the agreement, disagreement and neutral subsets $POS_{(\nu_{a},\nu_{r})}(X), NEG_{(\nu_{a},\nu_{r})}(X)$ and $BND_{(\nu_{a},\nu_{r})}(X)$ of $X$ as follows:
\begin{eqnarray*}
POS_{(\nu_{a},\nu_{r})}(X)
&=&\{x\in U\mid \nu_{a}(x,X)\in L^{+}_{a}\wedge \nu_{r}(x,X)\notin L^{-}_{r}\};\\
NEG_{(\nu_{a},\nu_{r})}(X)
&=&\{x\in U\mid \nu_{a}(x,X)\notin L^{+}_{a}\wedge \nu_{r}(x,X)\in L^{-}_{r}\};\\
BND_{(\nu_{a},\nu_{r})}(X)
&=&(POS_{(\nu_{a},\nu_{r})}(X)\cup NEG_{(\nu_{a},\nu_{r})}(X))^{C}\\
&=&\{x\in U\mid (\nu_{a}(x,X)\notin L^{+}_{a}\wedge \nu_{r}(x,X)\notin L^{-}_{r})\vee (\nu_{a}(x,X)\in L^{+}_{a}\wedge \nu_{r}(x,X)\in L^{-}_{r})\}.
\end{eqnarray*}
\end{definition}

For an agent $x\in U$ and a strategy $X\subseteq V$, if $\nu_{a}(x,X)\in L^{+}_{a}$ and $\nu_{r}(x,X)\notin L^{-}_{r}$, then $x\in POS_{(\nu_{a},\nu_{r})}(X)$; if $\nu_{a}(x,X)\notin L^{+}_{a}$ and $\nu_{r}(x,X)\in L^{-}_{r}$, then $x\in NEG_{(\nu_{a},\nu_{r})}(X)$; if $x\notin POS_{(\nu_{a},\nu_{r})}(X)$ and $x\notin NEG_{(\nu_{a},\nu_{r})}(X)$, then $x\in BND_{(\nu_{a},\nu_{r})}(X)$. That is, $POS_{(\nu_{a},\nu_{r})}(X)\cup NEG_{(\nu_{a},\nu_{r})}(X) \cup BND_{(\nu_{a},\nu_{r})}(X)=U$ and $POS_{(\nu_{a},\nu_{r})}(X)\cap NEG_{(\nu_{a},\nu_{r})}(X)=POS_{(\nu_{a},\nu_{r})}(X)\cap BND_{(\nu_{a},\nu_{r})}(X)=NEG_{(\nu_{a},\nu_{r})}(X)\cap BND_{(\nu_{a},\nu_{r})}(X)=\emptyset$. Furthermore, if the largest element $1\in L_{a}$, then $1\in L^{+}_{a}$; if the largest element $1\in L_{r}$, then $1\in L^{-}_{r}$; if $\nu_{a}(x,X)\preceq_{a}\nu_{a}(y,X)$ and $\nu_{a}(x,X)\in L^{+}_{a}$ for $x,y\in U$ and $X\subseteq V$, then $\nu_{a}(y,X)\in L^{+}_{a}$; if $\nu_{r}(x,X)\preceq_{r}\nu_{r}(y,X)$ and $\nu_{r}(x,X)\in L^{-}_{r}$ for $x,y\in U$ and $X\subseteq V$, then $\nu_{r}(y,X)\in L^{-}_{r}$. For simplicity, we denote $\ast^{C}$ as the complement of the set $\ast$ in the following.

Following Definition 3.2, we take the acceptance model $(A,\overline{A})$ and the rejection model $(R,\overline{R})$ to interpret $POS_{(\nu_{a},\nu_{r})}(X)$, $NEG_{(\nu_{a},\nu_{r})}(X)$ and $BND_{(\nu_{a},\nu_{r})}(X)$.
Firstly, we provide the acceptance region $POS_{\nu_{a}}(X)$ and the non-acceptance region $NPOS_{\nu_{a}}(X)$ of the strategy $X\subseteq V$ in $(A,\overline{A})$ as follows:
\begin{eqnarray*}
POS_{\nu_{a}}(X)
=\{x\in U\mid \nu_{a}(x,X)\in L^{+}_{a}\} \hspace{0.1cm}\text{ and }\hspace{0.1cm}
NPOS_{\nu_{a}}(X)
=\{x\in U\mid \nu_{a}(x,X)\notin L^{+}_{a}\}.
\end{eqnarray*}
For an agent $x\in U$, we get two-way decisions with respect to a strategy $X\subseteq V$ as follows:
$(A)$ if $\nu_{a}(x,X)\in L^{+}_{a}$, then we take an acceptance action, i.e. $x\in POS_{\nu_{a}}(X)$;
$(\overline{A})$ if $\nu_{a}(x,X)\notin L^{+}_{a}$, then we take a non-acceptance action, i.e. $x\in NPOS_{\nu_{a}}(X)$. The acceptance rule $(A)$ puts agents into an acceptance region, and the non-acceptance rule $(\overline{A})$ classifies agents into a non-acceptance region.
Secondly, we give the rejection region $NEG_{\nu_{a}}(X)$ and the non-rejection region $NNEG_{\nu_{a}}(X)$ for the strategy $X\subseteq V$ in $(R,\overline{R})$ as follows:
\begin{eqnarray*}
NEG_{\nu_{r}}(X)
=\{x\in U\mid \nu_{r}(x,X)\in L^{-}_{r}\} \hspace{0.1cm}\text{ and }\hspace{0.1cm}
NNEG_{\nu_{r}}(X)
=\{x\in U\mid \nu_{r}(x,X)\notin L^{-}_{r}\}.
\end{eqnarray*}
For an agent $x\in U$, we obtain two-way decisions with respect to a strategy $X\subseteq V$ as follows:
$(R)$ if $\nu_{r}(x,X)\in L^{-}_{r}$, then we take a rejection action, i.e. $x\in NEG_{\nu_{r}}(X)$;
$(\overline{R})$ if $\nu_{r}(x,X)\notin L^{-}_{r}$, then we take a non-rejection action, i.e. $x\in NNEG_{\nu_{r}}(X)$. The rejection rule $(R)$ classifies agents into a rejection region, and the non-rejection rule $(\overline{R})$ puts agents into a non-rejection region. Thirdly, we have the three-way decision rules for the agent $x\in U$ by combining $(A,\overline{A})$ and $(R,\overline{R})$ as follows:
$(P)$ if $\nu_{a}(x,X)\in L^{+}_{a}$ and $\nu_{r}(x,X)\notin L^{-}_{r}$, then $x\in POS_{(\nu_{a},\nu_{r})}(X)$;
$(R)$ if $\nu_{r}(x,X)\in L^{-}_{a}$ and $\nu_{a}(x,X)\notin L^{+}_{a}$, then $x\in NEG_{(\nu_{a},\nu_{r})}(X)$;
$(B)$ if $(\nu_{a}(x,X)\in L^{+}_{a}\wedge\nu_{r}(x,X)\in L^{-}_{r})$ or $(\nu_{a}(x,X)\notin L^{+}_{a}\wedge\nu_{r}(x,X)\notin L^{-}_{r})$, then $x\in BND_{(\nu_{a},\nu_{r})}(X)$.

It can be observed that an acceptance decision is interpreted as a combination of acceptance and non-rejection, i.e., $POS_{(\nu_{a},\nu_{r})}(X)=POS_{\nu_{a}}(X)\cap NNEG_{\nu_{r}}(X)$; a rejection decision is interpreted as a combination of rejection and non-acceptance, i.e., $NEG_{(\nu_{a},\nu_{r})}(X)= NEG_{\nu_{r}}(X)\cap NPOS_{\nu_{a}}(X)$; a non-commitment decision is interpreted as making an acceptance and a rejection decision simultaneously, or neither making an acceptance nor making a rejection decision, i.e., $BND_{(\nu_{a},\nu_{r})}(X)=(POS_{\nu_{a}}(X)\cap NEG_{\nu_{r}}(X))\cup (NPOS_{\nu_{a}}(X)\cap NNEG_{\nu_{r}}(X))$. Especially, we depict the above results by Table 2.

\begin{table}[H]\renewcommand{\arraystretch}{1.5}
\caption{Interpretations of $POS_{(\nu_{a},\nu_{r})}(X),NEG_{(\nu_{a},\nu_{r})}(X)$ and $BND_{(\nu_{a},\nu_{r})}(X)$.}
\tabcolsep0.44in
\begin{tabular}{ |c |c |c|}
\hline
 \diagbox{$(A,\overline{A})$}{$(R,\overline{R})$}&$\nu_{r}(x,X)\in L^{-}_{r}$ &$ \nu_{r}(x,X)\notin L^{-}_{r}$ \\\hline
\multirow{2}*{$\nu_{a}(x,X)\in L^{+}_{a}$}& $x\in BND_{(\nu_{a},\nu_{r})}(X)$ &$x\in POS_{(\nu_{a},\nu_{r})}(X)$  \\
&  (non-commitment) & (acceptance) \\\hline
\multirow{2}*{$\nu_{a}(x,X)\notin L^{+}_{a}$}& $x\in NEG_{(\nu_{a},\nu_{r})}(X)$ & $x\in BND_{(\nu_{a},\nu_{r})}(X)$\\
& (rejection) & (non-commitment)\\\hline
\end{tabular}
\end{table}

\subsection{SMZCAMs for Trisecting the Universe of Agents }

In this section, we investigate Sun, Ma and Zhao's conflict analysis model\cite{Sun1} for dividing agents into three disjoint blocks.

Suppose $U$ is the universe of agents, $V$ is the universe of issues, and $x\in U$, then the set-valued mapping $f=\{f^{+},f^{-}\}$ is defined as follows:
\begin{eqnarray*}
f^{+}: U\rightarrow P(V),\hspace{0.25cm}f^{+}(x)=\{c\in V\mid c(x)=+1\}\text{ and }
f^{-}: U\rightarrow P(V),\hspace{0.25cm}f^{-}(x)=\{c\in V\mid c(x)=-1\}.
\end{eqnarray*}

We see that the functions $f^{+}$ and $f^{-}$ are mapping from $U$ to $P(V)$, and
$f^{+}(x)$ stands for the issue subset of the universe $V$ which satisfies the agent $x$, and the image $f^{-}(x)$ denotes the issue subset of universe $V$ which are opposed by the agent $x$. Furthermore, we find that SMZCAMs employ the sets $\{f^{+}(x)\mid x\in U\}$ and $\{f^{-}(x)\mid x\in U\}$ to define the lower and upper approximations of a strategy $X\subseteq V$ as follows:
\begin{eqnarray*}
&&\underline{apr}^{+}_{f}(X)=\{x\in U\mid f^{+}(x)\subseteq X\},\hspace{1cm}\underline{apr}^{-}_{f}(X)=\{x\in U\mid f^{-}(x)\subseteq X\},\\
&&\overline{apr}^{+}_{f}(X)=\{x\in U\mid f^{+}(x)\cap X\neq \emptyset\},\hspace{0.4cm}\overline{apr}^{-}_{f}(X)=\{x\in U\mid f^{-}(x)\cap X\neq \emptyset\}.
\end{eqnarray*}

Obviously, we observe that the two lower approximations of sets are defined by using set inclusion $\subseteq$, and the two upper approximations of sets are defined by using set intersection $\cap$. Furthermore, we take two equivalence conditions $\neg(f^{+}(x)\subseteq X^{C})$ and
$\neg(f^{-}(x)\subseteq X^{C})$ to reconstruct the two upper approximations $\overline{apr}^{+}_{f}(X)$ and $\overline{apr}^{-}_{f}(X)$ of a strategy $X\subseteq V$ as follows:
\begin{eqnarray*}
\overline{apr}^{+}_{f}(X)=\{x\in U\mid \neg(f^{+}(x)\subseteq X^{C})\},\hspace{0.4cm}\overline{apr}^{-}_{f}(X)=\{x\in U\mid \neg(f^{-}(x)\subseteq X^{C})\}.
\end{eqnarray*}

\begin{definition}\cite{Sun1}
Let $U$ be the universe of agents, $V$ the universe of issues, and $X\subseteq V$ a strategy. Then the agreement, disagreement and neutral subsets $POS_{f}(X), NEG_{f}(X)$ and $BND_{f}(X)$ for $X$ are defined as follows:
\begin{eqnarray*}
&&POS_{f}(X)=\underline{apr}^{+}_{f}(X)-\underline{apr}^{-}_{f}(X)=\{x\in U\mid f^{+}(x)\subseteq X\}-\{x\in U\mid f^{-}(x)\subseteq X\};\\
&&NEG_{f}(X)=\underline{apr}^{-}_{f}(X)-\underline{apr}^{+}_{f}(X)=\{x\in U\mid f^{-}(x)\subseteq X\}-\{x\in U\mid f^{+}(x)\subseteq X\};\\
&&BND_{f}(X)=U-POS_{f}(X)\cup NEG_{f}(X).
\end{eqnarray*}
\end{definition}

We observe that the agreement subset $POS_{f}(X)$ is equal to the set $\{x\in U\mid f^{+}(x)\subseteq X\}$ subtracts the set $\{x\in U\mid f^{-}(x)\subseteq X\}$; the disagreement subset $NEG_{f}(X)$ is equal to the set $\{x\in U\mid f^{-}(x)\subseteq X\}$ subtracts the set $\{x\in U\mid f^{+}(x)\subseteq X\}$; the neutral subset $BND_{f}(X)$ is
equal to the universe $U$ of agents subtracts the union of $POS_{f}(X)$ and $NEG_{f}(X)$.
Furthermore, we find that
$POS_{f}(X)\cup NEG_{f}(X)\cup BND_{f}(X)=U,$ $POS_{f}(X)\cap NEG_{f}(X)=POS_{f}(X)\cap BND_{f}(X)=NEG_{f}(X)\cap BND_{f}(X)=\emptyset$. That is, we divide the universe of agents into three disjoint blocks with respect to $X$.

\begin{theorem}
Let $U$ be the universe of agents, $V$ the universe of issues, and $X\subseteq V$ a strategy. Then we have
\begin{eqnarray*}
&&POS_{f}(X)=\{x\in U\mid f^{+}(x)\subseteq X\}\cap\{x\in U\mid f^{-}(x)\cap X^{C}\neq \emptyset\};\\
&&NEG_{f}(X)=\{x\in U\mid f^{-}(x)\subseteq X\}\cap\{x\in U\mid f^{+}(x)\cap X^{C}\neq \emptyset\};\\
&&BND_{f}(X)=POS^{C}_{f}(X)\cap NEG^{C}_{f}(X).
\end{eqnarray*}
\end{theorem}

{\bf{Proof:}} Since $(\{x\in U\mid f^{-}(x)\subseteq X\})^{C}=\{x\in U\mid f^{-}(x)\cap X^{C}\neq \emptyset\}$,
by Definition 3.3, we have $POS_{f}(X)=\{x\in U\mid f^{+}(x)\subseteq X\}-\{x\in U\mid f^{-}(x)\subseteq X\}=\{x\in U\mid f^{+}(x)\subseteq X\}\cap (\{x\in U\mid f^{-}(x)\subseteq X\})^{C}=\{x\in U\mid f^{+}(x)\subseteq X\}\cap\{x\in U\mid f^{-}(x)\cap X^{C}\neq \emptyset\}.$ Then, since $(\{x\in U\mid f^{+}(x)\subseteq X\})^{C}=\{x\in U\mid f^{+}(x)\cap X^{C}\neq \emptyset\}$,
by Definition 3.3, we get $NEG_{f}(X)=\{x\in U\mid f^{-}(x)\subseteq X\}-\{x\in U\mid f^{+}(x)\subseteq X\}=\{x\in U\mid f^{-}(x)\subseteq X\}\cap (\{x\in U\mid f^{+}(x)\subseteq X\})^{C}=\{x\in U\mid f^{-}(x)\subseteq X\}\cap\{x\in U\mid f^{+}(x)\cap X^{C}\neq \emptyset\}.$ Finally, we have $BND_{f}(X)=U-POS_{f}(X)\cup NEG_{f}(X)=POS^{C}_{f}(X)\cap NEG^{C}_{f}(X).$ $\hspace{4.5cm}\Box$

Theorem 3.4 illustrates the agreement subset $POS_{f}(X)$ is equal to the intersection of $\{x\in U\mid f^{+}(x)\subseteq X\}$ and $\{x\in U\mid f^{-}(x)\cap X^{C}\neq \emptyset\}$, the disagreement subset $NEG_{f}(X)$ is equal to the intersection of $\{x\in U\mid f^{-}(x)\subseteq X\}$ and $\{x\in U\mid f^{+}(x)\cap X^{C}\neq \emptyset\}$, and the neutral subset $BND_{f}(X)$ is equal to the intersection of $POS^{C}_{f}(X)$ and $NEG^{C}_{f}(X)$.
Furthermore, for $x\in POS_{f}(X)$ and $y\in NEG_{f}(X)$, if $c_{i}(x)= +1$ and $c_{j}(y)=-1$ for $c_{i},c_{j}\in V$, then $c_{i}\in X$ and $ c_{j}\in X$; if $c_{i}(x)=-1$ for some $c_{i}\in X$, then there exists $c_{j}\in X^{C}$ such that $c_{j}(x)=-1$;
if $c_{i}(y)=+1$ for some $c_{i}\in X$, then there exists $c_{j}\in X^{C}$ such that $c_{j}(y)=+1$.

\begin{theorem}
Let $U$ be the universe of agents, $V$ the universe of issues, and $X\subseteq V$ a strategy. Then we have
\begin{eqnarray*}
POS_{f}(X)&=&\{x\in U\mid f^{+}(x)\subseteq X\}\cap\{x\in U\mid \neg(f^{-}(x)\subseteq X)\};\\
NEG_{f}(X)&=&\{x\in U\mid f^{-}(x)\subseteq X\}\cap\{x\in U\mid \neg(f^{+}(x)\subseteq X)\};\\
BND_{f}(X)&=&(\{x\in U\mid f^{+}(x)\subseteq X\}\cap\{x\in U\mid f^{-}(x)\subseteq X\})\cup (\{x\in U\mid \neg(f^{+}(x)\subseteq X)\}\cap\{x\in U\mid \neg(f^{-}(x)\subseteq X)\}).
\end{eqnarray*}
\end{theorem}

{\bf{Proof:}} It is straightforward by Theorem 3.4.$ \hspace{8.5cm}\Box$

%
%

Theorem 3.5 illustrates that the agreement subset $POS_{f}(X)$, disagreement subset $NEG_{f}(X)$ and neutral subset $BND_{f}(X)$ of the strategy $X$ are defined uniformly by using the set inclusion. Furthermore, we study the relationship between Definition 3.2 and Definition 3.3 as follows.

\begin{definition}
Let $U$ be the universe of agents, $V$ the universe of issues, $L_{a}=L_{r}=\{T,F\}$ with $F\preceq T$, $L^{+}_{a}=L^{-}_{r}=\{T\}$, $x\in U$, and $X\subseteq V$ a strategy. Then the acceptance evaluation $\nu_{a}(x,X)$ and the rejection evaluation $\nu_{r}(x,X)$ are defined as follows:
$$\nu_{a}(x,X)=\left\{
\begin{array}{ccc}
T,&{\rm if}& f^{+}(x)\subseteq X,\\
F,&{\rm if}& \neg(f^{+}(x)\subseteq X),\\
\end{array}
\right. \hspace{0.1cm}\text{ and }\hspace{0.2cm} \nu_{r}(x,X)=\left\{
\begin{array}{ccc}
T,&{\rm if}& f^{-}(x)\subseteq X^{C},\\
F,&{\rm if}& \neg(f^{-}(x)\subseteq X^{C}).\\
\end{array}
\right.$$
\end{definition}

We observe that $\nu_{a}(x,X)$ and $\nu_{r}(x,X)$ take the value from the set $\{T,F\}$. Concretely, if $f^{+}(x)\subseteq X,$ then $\nu_{a}(x,X)=T$; if $f^{+}(x)\nsubseteq X,$ then $\nu_{a}(x,X)=F$; if $f^{-}(x)\subseteq X,$ then $\nu_{r}(x,X)=T$; if $f^{-}(x)\nsubseteq X,$ then $\nu_{r}(x,X)=F$.
Furthermore, by Theorem 3.5 and Definition 3.6, we reconstruct the agreement, disagreement and neutral subsets given by Definition 3.3 as follows:
\begin{eqnarray*}
POS_{f}(X)=POS_{(\nu_{a},\nu_{r})}(X)
&=&\{x\in U\mid \nu_{a}(x,X)\in L^{+}_{a}\wedge \nu_{r}(x,X)\notin L^{-}_{r}\};\\
NEG_{f}(X)=NEG_{(\nu_{a},\nu_{r})}(X)
&=&\{x\in U\mid \nu_{a}(x,X)\notin L^{+}_{a}\wedge \nu_{r}(x,X)\in L^{-}_{r}\};\\
BND_{f}(X)=BND_{(\nu_{a},\nu_{r})}(X)
&=&(POS_{(\nu_{a},\nu_{r})}(X)\cup NEG_{(\nu_{a},\nu_{r})}(X))^{C}\\
&=&\{x\in U\mid (\nu_{a}(x,X)\notin L^{+}_{a}\wedge \nu_{r}(x,X)\notin L^{-}_{r})\vee (\nu_{a}(x,X)\in L^{+}_{a}\wedge \nu_{r}(x,X)\in L^{-}_{r})\}.
\end{eqnarray*}

Belows, we employ the acceptance model $(A,\overline{A})$ and the rejection model $(R,\overline{R})$ to interpret $POS_{f}(X)$, $NEG_{f}(X)$ and $BND_{f}(X)$.
Firstly, we provide the acceptance region $POS_{f^{+}}(X)$ and the non-acceptance region $NPOS_{f^{+}}(X)$ of the strategy $X\subseteq V$ in  $(A,\overline{A})$ as follows:
\begin{eqnarray*}
POS_{f^{+}}(X)
=\{x\in U\mid f^{+}(x)\subseteq X\} \hspace{0.1cm}\text{ and }\hspace{0.1cm}
NPOS_{f^{+}}(X)
=\{x\in U\mid f^{+}(x)\nsubseteq X\}.
\end{eqnarray*}
For an agent $x\in U$, we have two-way decisions with respect to a strategy $X\subseteq V$ as follows:
$(A)$ if $f^{+}(x)\subseteq X$, then we take an acceptance action, i.e. $x\in POS_{f^{+}}(X)$;
$(\overline{A})$ if $f^{+}(x)\nsubseteq X$, then we take a non-acceptance action, i.e. $x\in NPOS_{f^{+}}(X)$. The acceptance rule $(A)$ classifies agents into an acceptance region, and the non-acceptance rule $(\overline{A})$ classifies agents into a non-acceptance region. Secondly, we provide the rejection region $NEG_{f^{-}}(X)$ and the non-rejection region $NNEG_{f^{-}}(X)$ of the strategy $X\subseteq V$ in $(R,\overline{R})$ as follows:
\begin{eqnarray*}
NEG_{f^{-}}(X)
=\{x\in U\mid f^{-}(x)\subseteq X\} \hspace{0.1cm}\text{ and }\hspace{0.1cm}
NNEG_{f^{-}}(X)
=\{x\in U\mid f^{-}(x)\nsubseteq X\}.
\end{eqnarray*}
For an agent $x\in U$, we take two-way decisions with respect to a strategy $X\subseteq V$ as follows:
$(R)$ if $f^{-}(x)\subseteq X$, then we take a rejection action, i.e. $x\in NEG_{f^{-}}(X)$;
$(\overline{R})$ if $f^{-}(x)\nsubseteq X$, then we take a non-rejection action, i.e. $x\in NNEG_{f^{-}}(X)$. The rejection rule $(R)$ puts agents into the rejection region, and the non-rejection rule $(\overline{R})$ classifies agents into the non-rejection region. Thirdly, we have the three-way decision rules for the agent $x\in U$ by combining $(A,\overline{A})$ and $(R,\overline{R})$ as follows:
$(P)$ if $f^{+}(x)\subseteq X$ and $f^{-}(x)\nsubseteq X$, then $x\in POS_{f}(X)$;
$(R)$ if $f^{-}(x)\subseteq X$ and $f^{+}(x)\nsubseteq X$, then $x\in NEG_{f}(X)$;
$(B)$ if $(f^{+}(x)\subseteq X\wedge f^{-}(x)\subseteq X)$ or $(f^{+}(x)\nsubseteq X\wedge f^{-}(x)\nsubseteq X)$, then $x\in BND_{f}(X)$.

We see that $POS_{f}(X)=POS_{f^{+}}(X)\cap NNEG_{f^{-}}(X)$, $NEG_{f}(X)=NPOS_{f^{+}}(X)\cap NEG_{f^{-}}(X)$ and $BND_{f}(X)=(POS_{f^{+}}(X)\cap NEG_{f^{-}}(X))\cup (NPOS_{f^{+}}(X)\cap NNEG_{f^{-}}(X))$. Meanwhile, we interpret the three-way decision rules for the agent $x\in U$ by Table 3 and illustrate that SMZCAs are special cases of TWDCAMs given by Definition 3.2.

\begin{table}[H]\renewcommand{\arraystretch}{1.5}
\caption{Interpretations of $POS_{f}(X),NEG_{f}(X)$ and $BND_{f}(X)$.}
\tabcolsep0.41in
\begin{tabular}{ |c |c |c|}
\hline
 \diagbox{$(A,\overline{A})$}{$(R,\overline{R})$}&$\nu_{r}(x,X)=T\in L^{-}_{r}$ &$ \nu_{r}(x,X)=F\notin L^{-}_{r}$ \\\hline
\multirow{2}*{$\nu_{a}(x,X)=T\in L^{+}_{a}$}&  $x\in BND_{f}(X)$ &$x\in POS_{f}(X)$  \\
& (non-commitment) & (acceptance) \\\hline
\multirow{2}*{$\nu_{a}(x,X)=F\notin L^{+}_{a}$}&$x\in NEG_{f}(X)$ &$x\in BND_{f}(X)$\\
& (rejection) & (non-commitment)\\\hline
\end{tabular}
\end{table}

\subsection{FQWCAMs for Trisecting the Universe of Agents}

In this section, we investigate Fan, Qi and Wei's conflict analysis model\cite{Fan} for trisecting the universe of agents.

\begin{definition}\cite{Fan}
Let $U$ be the universe of agents, $V$ the universe of issues, $([0,1],\leq)$ a totally ordered set, $X\subseteq V$ a strategy, $D(f^{+}(x)\mid X)=\frac{|f^{+}(x)\cap X|}{|X|}$ and $
D(f^{-}(x)\mid X)=\frac{|f^{-}(x)\cap X|}{|X|}$. Then the functions $\mu_{a}$ and $\mu_{r}$ are defined as follows:
\begin{eqnarray*}
&&\mu_{a}: U\times P(V)\longrightarrow [0,1],\hspace{0.25cm}\mu_{a}(x,X)=D(f^{+}(x)\mid X)=\frac{|f^{+}(x)\cap X|}{|X|};\\
&&\mu_{r}: U\times P(V)\longrightarrow [0,1],\hspace{0.25cm}\mu_{r}(x,X)=D(f^{-}(x)\mid X)=\frac{|f^{-}(x)\cap X|}{|X|}.
\end{eqnarray*}
\end{definition}

The functions $\mu_{a}$ and $\mu_{r}$ are mappings from $U\times P(V)$ to $[0,1]$, and $\mu_{a}(x,X)$ evaluates the extent to which the agent $x$ accepts the strategy $X$ and $\mu_{r}(x,X)$ evaluates the extent to which the agent $x$ rejects the strategy $X$. Furthermore, if we take $L_{a}=L_{r}=[0,1]$ and $\preceq_{a}=\preceq_{r}=\leq$, then $\mu_{a}$ and $\mu_{r}$ are special acceptance and rejection evaluations given by Definition 3.1, respectively.

\begin{definition}\cite{Fan}
Let $U$ be the universe of agents, $V$ the universe of issues, $(\alpha,1]$ the designated values for acceptance, $(\beta,1]$ the designated values for rejection, and $X\subseteq V$ a strategy. Then we define the $(\alpha,\beta)-$agreement, $(\alpha,\beta)-$disagreement and $(\alpha,\beta)-$neutral subsets $POS_{(\mu_{a},\mu_{r})}(X)$, $NEG_{(\mu_{a},\mu_{r})}(X)$ and $BND_{(\mu_{a},\mu_{r})}(X)$ of $X$ as follows:
\begin{eqnarray*}
&&POS_{(\mu_{a},\mu_{r})}(X)=\{x\in U\mid \mu_{a}(x,X)\in (\alpha,1]\wedge \mu_{r}(x,X)\notin (\beta,1]\};\\
&&NEG_{(\mu_{a},\mu_{r})}(X)=\{x\in U\mid \mu_{a}(x,X)\notin (\alpha,1]\wedge \mu_{r}(x,X)\in (\beta,1]\};\\
&&BND_{(\mu_{a},\mu_{r})}(X)=U-POS_{(\alpha,\beta)}(X)\cup NEG_{(\alpha,\beta)}(X).
\end{eqnarray*}
\end{definition}

We employ two functions $\mu_{a}(x,X)$ and $\mu_{r}(x,X)$ to trisect the universe of agents into $POS_{(\mu_{a},\mu_{r})}(X)$, $NEG_{(\mu_{a},\mu_{r})}(X)$ and $BND_{(\mu_{a},\mu_{r})}(X)$ with respect to $X$. Furthermore, if
$L^{+}_{a}=(\alpha,1]$, $L^{+}_{r}=(\beta,1]$, $\nu_{a}(x,X)=\mu_{a}(x,X)=\frac{|f^{+}(x)\cap X|}{|X|}$, and $\nu_{r}(x,X)=\mu_{r}(x,X)=\frac{|f^{-}(x)\cap X|}{|X|}$, then we have
\begin{eqnarray*}
&&POS_{(\mu_{a},\mu_{r})}(X)=POS_{(\nu_{a},\nu_{r})}(X)=\{x\in U\mid \nu_{a}(x,X)\in L^{+}_{a}\wedge \nu_{r}(x,X)\notin L^{-}_{a}\};\\
&&NEG_{(\mu_{a},\mu_{r})}(X)=NEG_{(\nu_{a},\nu_{r})}(X)=\{x\in U\mid \nu_{a}(x,X)\notin L^{+}_{a}\wedge \nu_{r}(x,X)\in L^{-}_{a}\};\\
&&BND_{(\mu_{a},\mu_{r})}(X)=BND_{(\nu_{a},\nu_{r})}(X)
=\{x\in U\mid (\nu_{a}(x,X)\notin L^{+}_{a}\wedge \nu_{r}(x,X)\notin L^{-}_{r})\vee (\nu_{a}(x,X)\in L^{+}_{a}\wedge \nu_{r}(x,X)\in L^{-}_{r})\}.
\end{eqnarray*}

Following Definition 3.8, we interpret $POS_{(\mu_{a},\mu_{r})}(X)$, $NEG_{(\mu_{a},\mu_{r})}(X)$ and $BND_{(\mu_{a},\mu_{r})}(X)$ with the acceptance model $(A,\overline{A})$ and the rejection model $(R,\overline{R})$ as follows. Firstly,
we provide the acceptance region $POS_{\mu_{a}}(X)$ and the non-acceptance region  $NPOS_{\mu_{a}}(X)$ of the strategy $X\subseteq V$ in $(A,\overline{A})$ as follows:
\begin{eqnarray*}
POS_{\mu_{a}}(X)
=\{x\in U\mid \mu_{a}(x,X)\in (\alpha,1]\} \hspace{0.1cm}\text{ and }\hspace{0.1cm}
NPOS_{\mu_{a}}(X)
=\{x\in U\mid \mu_{a}(x,X)\notin (\alpha,1]\}.
\end{eqnarray*}
For an agent $x\in U$, we make two-way decisions with respect to a strategy $X$ as follows:
$(A)$ if $\mu_{a}(x,X)\in (\alpha,1]$, then we take an acceptance action, i.e. $x\in POS_{\mu_{a}}(X)$;
$(\overline{A})$ if $\mu_{a}(x,X)\notin (\alpha,1]$, then we take a non-acceptance action, i.e. $x\in NPOS_{\mu_{a}}(X)$. The acceptance rule $(A)$ classifies agents into an acceptance region, and the non-acceptance rule $(\overline{A})$ classifies agents into a non-acceptance region.
Secondly, we provide the rejection region $NEG_{\mu_{a}}(X)$ and the non-rejection region $NNEG_{\mu_{a}}(X)$ of the strategy $X\subseteq V$ in $(R,\overline{R})$ as follows:
\begin{eqnarray*}
NEG_{\mu_{r}}(X)
=\{x\in U\mid \mu_{r}(x,X)\in (\beta,1]\} \hspace{0.1cm}\text{ and }\hspace{0.1cm}
NNEG_{\mu_{r}}(X)
=\{x\in U\mid \mu_{r}(x,X)\notin (\beta,1]\}.
\end{eqnarray*}
For an agent $x\in U$, we derive two-way decisions with respect to a strategy $X$ as follows:
$(A)$ if $\mu_{r}(x,X)\in (\beta,1]$, then we take a rejection action, i.e. $x\in NEG_{\mu_{r}}(X)$;
$(\overline{A})$ if $\mu_{r}(x,X)\notin (\beta,1]$, then we take a non-rejection action, i.e. $x\in NNEG_{\mu_{r}}(X)$. The rejection rule $(R)$ classifies agents into a rejection region, and the non-rejection rule $(\overline{R})$ puts agents into a non-rejection region. Thirdly, we have the three-way decision rules for the agent $x\in U$ by combining $(A,\overline{A})$ and $(R,\overline{R})$ as follows:
$(P)$ if $\mu_{a}(x,X)\in (\alpha,1]$ and $\mu_{r}(x,X)\notin (\beta,1]$, then $x\in POS_{(\mu_{a},\mu_{r})}(X)$;
$(R)$ if $\mu_{r}(x,X)\in (\beta,1]$ and $\mu_{a}(x,X)\notin (\alpha,1]$, then $x\in NEG_{(\mu_{a},\mu_{r})}(X)$;
$(B)$ if $(\mu_{a}(x,X)\in (\alpha,1]\wedge\mu_{r}(x,X)\in (\beta,1])$ or $(\mu_{a}(x,X)\notin (\alpha,1]\wedge\mu_{r}(x,X)\notin (\beta,1])$, then $x\in BND_{(\mu_{a},\mu_{r})}(X)$.

We see that $POS_{(\mu_{a},\mu_{r})}(X)=POS_{\mu_{a}}(X)\cap NNEG_{\mu_{r}}(X)$, $NEG_{(\mu_{a},\mu_{r})}(X)=NPOS_{\mu_{a}}(X)\cap NEG_{\mu_{r}}(X)$ and $BND_{(\mu_{a},\mu_{r})}(X)=(POS_{\mu_{a}}(X)\cap NEG_{\mu_{r}}(X))\cup (NPOS_{\mu_{a}}(X)\cap NNEG_{\mu_{r}}(X))$. Furthermore, we depict the three-way decision rules for the agent $x\in U$ by Table 4 and illustrate that FQWCAMs are special cases of TWDCAMs given by Definition 3.2.

\begin{table}[H]\renewcommand{\arraystretch}{1.5}
\caption{Interpretations of $POS_{(\mu_{a},\mu_{r})}(X),NEG_{(\mu_{a},\mu_{r})}(X)$ and $BND_{(\mu_{a},\mu_{r})}(X)$.}
\tabcolsep0.35in
\begin{tabular}{ |c |c |c|}
\hline
 \diagbox{$(A,\overline{A})$}{$(R,\overline{R})$}&$\mu_{r}(x,X)\in L^{-}_{r}=(\beta,1]$ &$ \mu_{r}(x,X)\notin L^{-}_{r}=(\beta,1]$ \\\hline
\multirow{2}*{$\mu_{a}(x,X)\in L^{+}_{a}=(\alpha,1]$}& $x\in BND_{(\mu_{a},\mu_{r})}(X)$ &$x\in POS_{(\mu_{a},\mu_{r})}(X)$  \\
& (non-commitment) & (acceptance) \\\hline
\multirow{2}*{$\mu_{a}(x,X)\notin L^{+}_{a}=(\alpha,1]$}& $x\in NEG_{(\mu_{a},\mu_{r})}(X)$ &$x\in BND_{(\mu_{a},\mu_{r})}(X)$\\
& (rejection) & (non-commitment)\\\hline
\end{tabular}
\end{table}

\section{TWDCAMs for Trisecting the Universe of Issues}

In this section, we study TWDCAMs for dividing issue sets into three disjoint blocks.

\subsection{TWDCAMs II}

In this section, we provide another type of TWDCAMs for trisecting the universe of issues.

\begin{definition}
Let $U$ be the universe of agents, and $V$ the universe of issues, $(L_{a},\preceq_{a})$ and $(L_{r},\preceq_{r})$
are two posets. Then a pair of functions $\omega_{a}: V\times P(U)\rightarrow L_{a}$ and $\omega_{r}: V\times P(U)\rightarrow L_{r}$ is called an acceptance evaluation and a rejection evaluation of an issue by an agent group, respectively.
\end{definition}

For an issue $c\in V$ and an agent group $Y\subseteq U$, we refer to $\omega_{a}(c,Y)$ and $\omega_{r}(c,Y)$ as the acceptance and rejection degrees of the issue $c$ by the agent group $Y$, respectively. For $c_{i},c_{j}\in V$, if $\omega_{a}(c_{i},Y)\preceq_{a} \omega_{a}(c_{j},Y)$, then $c_{i}$ is less acceptable than $c_{j}$ by $Y$; if $\omega_{r}(c_{i},Y)\preceq_{r} \omega_{r}(c_{j},Y)$, then $c_{i}$ is less rejectable than $c_{j}$ by $Y$. Furthermore, if an issue $c$ is accepted by an agent group $Y$, then $\omega_{a}(c,Y)$ must be in a certain subset $L^{+}_{a}\subseteq L_{a}$, where $L^{+}_{a}$ denotes the acceptance region of $L_{a}$; if an issue $c$ is rejected by an agent group $Y$, then $\omega_{r}(c,Y)$ must be in a certain subset $L^{-}_{r}\subseteq L_{r}$, where $L^{-}_{r}$ stands for the rejection region of $L_{r}$. For simplicity, $L^{+}_{a}$ and $L^{-}_{r}$ are called the designated values for acceptance and designed values for rejection, respectively.

\begin{definition}
Let $U$ be the universe of agents, $V$ the universe of issues, $L^{+}_{a}$ and $L^{-}_{r}$ the designated values for acceptance and designed values for rejection, respectively, $\omega_{a}$ and $\omega_{r}$ the acceptance evaluation and rejection evaluation of an issue by an agent group, respectively, and $Y\subseteq U$ an agent group. Then we define the agreement, disagreement and neutral strategies $POS_{(\omega_{a},\omega_{r})}(Y),NEG_{(\omega_{a},\omega_{r})}(Y)$ and $BND_{(\omega_{a},\omega_{r})}(Y)$ of $Y$ as follows:
\begin{eqnarray*}
POS_{(\omega_{a},\omega_{r})}(Y)
&=&\{c\in V\mid \omega_{a}(c,Y)\in L^{+}_{a}\wedge \omega_{r}(c,Y)\notin L^{-}_{r}\};\\
NEG_{(\omega_{a},\omega_{r})}(Y)
&=&\{c\in V\mid \omega_{a}(c,Y)\notin L^{+}_{a}\wedge \omega_{r}(c,Y)\in L^{-}_{r}\};\\
BND_{(\omega_{a},\omega_{r})}(Y)
&=&(POS_{(\omega_{a},\omega_{r})}(Y)\cup NEG_{(\omega_{a},\omega_{r})}(Y))^{C}\\
&=&\{c\in V\mid (\omega_{a}(c,Y)\notin L^{+}_{a}\wedge \omega_{r}(c,Y)\notin L^{-}_{r})\vee (\omega_{a}(c,Y)\in L^{+}_{a}\wedge \omega_{r}(c,Y)\in L^{-}_{r})\}.
\end{eqnarray*}
\end{definition}

For an issue $c\in V$ and an agent group $Y\subseteq U$, if $\omega_{a}(c,Y)\in L^{+}_{a}$ and $\omega_{r}(c,Y)\notin L^{-}_{r}$, then $c\in POS_{(\omega_{a},\omega_{r})}(Y)$; if $\omega_{a}(c,Y)\in L^{-}_{a}$ and $\omega_{r}(c,Y)\notin L^{+}_{r}$, then $c\in NEG_{(\omega_{a},\omega_{r})}(Y)$; if $c\notin POS_{(\omega_{a},\omega_{r})}(Y)$ and $c\notin NEG_{(\omega_{a},\omega_{r})}(Y)$, then $c\in BND_{(\omega_{a},\omega_{r})}(Y)$. That is, $POS_{(\omega_{a},\omega_{r})}(Y)\cup NEG_{(\omega_{a},\omega_{r})}(Y) \cup BND_{(\omega_{a},\omega_{r})}(Y)=U$ and $POS_{(\omega_{a},\omega_{r})}(Y)\cap NEG_{(\omega_{a},\omega_{r})}(Y)=POS_{(\omega_{a},\omega_{r})}(Y)\cap BND_{(\omega_{a},\omega_{r})}(Y)=NEG_{(\omega_{a},\omega_{r})}(Y)\cap BND_{(\omega_{a},\omega_{r})}(Y)=\emptyset$. Furthermore, if the largest element $1\in L_{a}$, then $1\in L^{+}_{a}$; if the largest element $1\in L_{r}$, then $1\in L^{-}_{r}$; if $\omega_{a}(c_{i},Y)\preceq_{a}\omega_{a}(c_{j},Y)$ and $\omega_{a}(c_{i},Y)\in L^{+}_{a}$ for $c_{i},c_{j}\in U$ and $Y\subseteq U$, then $\omega_{a}(c_{j},Y)\in L^{+}_{a}$; if $\omega_{r}(c_{i},Y)\preceq_{r}\omega_{r}(c_{j},Y)$ and $\omega_{r}(c_{i},Y)\in L^{-}_{r}$ for $c_{i},c_{j}\in V$ and $Y\subseteq U$, then $\omega_{r}(c_{j},Y)\in L^{-}_{r}$.

Following Definition 4.2, we employ the acceptance model $(A^{\ast},\overline{A^{\ast}})$ and the rejection model $(R^{\ast},\overline{R^{\ast}})$ to interpret $POS_{(\omega_{a},\omega_{r})}(Y)$, $NEG_{(\omega_{a},\omega_{r})}(Y)$ and $BND_{(\omega_{a},\omega_{r})}(Y)$ as follows. Firstly, we provide the acceptance region $POS_{\omega_{a}}(Y)$ and the non-acceptance region $NPOS_{\omega_{a}}(Y)$ of the agent group $Y\subseteq U$ in $(A^{\ast},\overline{A^{\ast}})$ as follows:
\begin{eqnarray*}
POS_{\omega_{a}}(Y)
=\{c\in V\mid \omega_{a}(c,Y)\in L^{+}_{a}\} \hspace{0.1cm}\text{ and }\hspace{0.1cm}
NPOS_{\omega_{a}}(Y)
=\{c\in V\mid \omega_{a}(c,Y)\notin L^{+}_{a}\}.
\end{eqnarray*}
For an issue $c\in V$, we make two-way decisions with respect to an agent group $Y$ as follows:
$(A^{\ast})$ if $\omega_{a}(c,Y)\in L^{+}_{a}$, then we take an acceptance action, i.e. $c\in POS_{\omega_{a}}(Y)$;
$(\overline{A^{\ast}})$ if $\omega_{a}(c,Y)\notin L^{+}_{a}$, then we take a non-acceptance action, i.e. $c\in NPOS_{\omega_{a}}(Y)$. The acceptance rule $(A^{\ast})$ puts issues into an acceptance region, and the non-acceptance rule $(\overline{A^{\ast}})$ classifies issues into a non-acceptance region.
Secondly, we give the rejection region $NEG_{\omega_{a}}(Y)$ and the non-rejection region $NNEG_{\omega_{a}}(Y)$ of the agent group $Y\subseteq V$ in $(R^{\ast},\overline{R^{\ast}})$ as follows:
\begin{eqnarray*}
NEG_{\omega_{r}}(Y)
=\{c\in V\mid \omega_{r}(c,Y)\in L^{-}_{r}\} \hspace{0.1cm}\text{ and }\hspace{0.1cm}
NNEG_{\omega_{r}}(Y)
=\{c\in V\mid \omega_{r}(c,Y)\notin L^{-}_{r}\}.
\end{eqnarray*}
For an issue $c\in V$, we make two-way decisions with respect to an agent group $Y$ as follows:
$(A^{\ast})$ if $\omega_{r}(c,Y)\in L^{-}_{r}$, then we take a rejection action, i.e. $x\in NEG_{\omega_{r}}(Y)$;
$(\overline{A^{\ast}})$ if $\omega_{r}(c,Y)\notin L^{-}_{r}$, then we take a non-rejection action, i.e. $x\in NNEG_{\omega_{r}}(Y)$. The rejection rule $(R^{\ast})$ classifies issues into a rejection region, and the non-rejection rule $(\overline{R^{\ast}})$ puts issues into a non-rejection region. Thirdly, we have the three-way decision rules for the issue $c\in V$ by combining $(A^{\ast},\overline{A^{\ast}})$ and $(R^{\ast},\overline{R^{\ast}})$ as follows:
$(P)$ if $\omega_{a}(c,Y)\in L^{+}_{a}$ and $\omega_{r}(c,Y)\notin L^{-}_{r}$, then $c\in POS_{(\omega_{a},\omega_{r})}(Y)$;
$(R)$ if $\omega_{r}(c,Y)\in L^{-}_{a}$ and $\omega_{a}(c,Y)\notin L^{+}_{a}$, then $c\in NEG_{(\omega_{a},\omega_{r})}(Y)$;
$(B)$ if $(\omega_{a}(c,Y)\in L^{+}_{a}\wedge\omega_{r}(c,Y)\in L^{-}_{r})$ or $(\omega_{a}(c,Y)\notin L^{+}_{a}\wedge\omega_{r}(c,Y)\notin L^{-}_{r})$, then $c\in BND_{(\omega_{a},\omega_{r})}(Y)$.

\begin{table}[H]\renewcommand{\arraystretch}{1.5}
\caption{Interpretations of $POS_{(\omega_{a},\omega_{r})}(Y),NEG_{(\omega_{a},\omega_{r})}(Y)$ and $BND_{(\omega_{a},\omega_{r})}(Y)$.}
\tabcolsep0.381in
\begin{tabular}{ |c |c |c|}
\hline
 \diagbox{$(A^{\ast},\overline{A^{\ast}})$}{$(R^{\ast},\overline{R^{\ast}})$}&$\omega_{a}(c,Y)\in L^{-}_{r}$ &$ \nu_{a}(c,Y)\notin L^{-}_{r}$ \\\hline
\multirow{2}*{$\omega_{a}(c,Y)\in L^{+}_{a}$}& $c\in BND_{(\omega_{a},\omega_{r})}(Y)$ &$c\in POS_{(\omega_{a},\omega_{r})}(Y)$  \\
& (non-commitment) & (acceptance) \\\hline
\multirow{2}*{$\omega_{a}(c,Y)\notin L^{+}_{a}$} & $c\in NEG_{(\omega_{a},\omega_{r})}(Y)$  &$c\in BND_{(\omega_{a},\omega_{r})}(Y)$\\
&(rejection) & (non-commitment)\\\hline
\end{tabular}
\end{table}

We observe that $POS_{(\omega_{a},\omega_{r})}(Y)=POS_{\omega_{a}}(Y)\cap NNEG_{\omega_{r}}(Y)$, $NEG_{(\omega_{a},\omega_{r})}(Y)=NPOS_{\omega_{a}}(Y)\cap NEG_{\omega_{r}}(Y)$ and $BND_{(\omega_{a},\omega_{r})}(Y)=(POS_{\omega_{a}}(Y)\cap NEG_{\omega_{r}}(Y))\cup (NPOS_{\omega_{a}}(Y)\cap NNEG_{\omega_{r}}(Y))$.
Furthermore, we depict the three-way decision rules for the issue $c\in V$ by Table 5.

\subsection{SMZCAMs for Trisecting the Universe of Issues}

In this section, we investigate Sun, Ma and Zhao's conflict analysis model\cite{Sun1}
for dividing issues into three disjoint blocks.

Suppose $U$ is the universe of agents, $V$ is the universe of issues, $c\in V$, and $Y\subseteq U$ is an agent group, then the set-valued mappings $g^{+}$ and $g^{-}$ is defined as follows:
\begin{eqnarray*}
g^{+}: V\rightarrow P(U),\hspace{0.25cm}g^{+}(c)=\{x\in U\mid c(x)=+1\} \text{ and }
g^{-}: V\rightarrow P(U),\hspace{0.25cm}g^{-}(c)=\{x\in U\mid c(x)=-1\}.
\end{eqnarray*}

The functions $g^{+}$ and $g^{-}$ are mapping from $V$ to $P(U)$, and
$g^{+}(c)$ represents the agent subset of the universe $U$ who support the issue $c$, and the image $g^{-}(c)$ denotes the agent subset of universe $U$ who oppose  the issue $c$.
Furthermore, we observe that SMZCAMs employ the sets $\{g^{+}(c)\mid c\in V\}$ and $\{g^{-}(c)\mid c\in V\}$ to define the lower and upper approximations of an agent group $Y$ as follows:
\begin{eqnarray*}
&&\underline{apr}^{+}_{g}(Y)=\{c\in V\mid g^{+}(c)\subseteq Y\},\hspace{1cm}\underline{apr}^{-}_{g}(Y)=\{c\in V\mid g^{-}(c)\subseteq Y\},\\
&&\overline{apr}^{+}_{g}(Y)=\{c\in V\mid g^{+}(c)\cap Y\neq \emptyset\},\hspace{0.4cm}\overline{apr}^{+}_{g}(Y)=\{c\in V\mid g^{-}(c)\cap Y\neq \emptyset\}.
\end{eqnarray*}

Obviously, we see that the two lower approximations of sets are defined by using the set inclusion $\subseteq$, and the two upper approximations of sets are defined by using the set intersection $\cap$. Furthermore, we take two equivalence conditions $\neg(g^{+}(c)\subseteq Y^{C})$ and
$\neg(g^{-}(c)\subseteq Y^{C})$ to reconstruct the upper approximations of an agent group $Y$ as follows:
\begin{eqnarray*}
\overline{apr}^{+}_{g}(Y)=\{c\in V\mid \neg(g^{+}(c)\subseteq Y^{C})\},\hspace{0.4cm}\overline{apr}^{-}_{g}(Y)=\{c\in V\mid \neg(g^{-}(c)\subseteq Y^{C})\}.
\end{eqnarray*}

\begin{definition}\cite{Sun1}
Let $U$ be the universe of agents, $V$ the universe of issues, and $Y\subseteq U$ an agent group. Then the agreement, disagreement and neutral strategies $POS_{g}(Y)$, $NEG_{g}(Y)$ and $BND_{g}(Y)$ of $Y$ are defined as follows:
\begin{eqnarray*}
&&POS_{g}(Y)=\{c\in V\mid g^{+}(c)\subseteq Y\}-\{c\in V\mid g^{-}(c)\subseteq Y\};\\
&&NEG_{g}(Y)=\{c\in V\mid g^{-}(c)\subseteq Y\}-\{c\in V\mid g^{+}(c)\subseteq Y\};\\
&&BND_{g}(Y)=V-POS_{g}(Y)\cup NEG_{g}(Y).
\end{eqnarray*}
\end{definition}

We observe that the agreement strategy $POS_{g}(Y)$ is equal to the set $\{c\in V\mid g^{+}(c)\subseteq Y\}$ subtracts the set $\{c\in V\mid g^{-}(x)\subseteq X\}$; the disagreement strategy $NEG_{f}(Y)$ is equal to $\{c\in V\mid g^{-}(c)\subseteq Y\}$ subtracts the set $\{c\in V\mid g^{+}(c)\subseteq Y\}$; the neutral strategy $BND_{g}(Y)$ is
equal to the universe $V$ of issues subtracts the union of $POS_{g}(Y)$ and $NEG_{g}(Y)$.
Furthermore, we have $POS_{g}(Y)\cup NEG_{g}(Y)\cup BND_{g}(Y)=V$ and $POS_{g}(Y)\cap NEG_{g}(Y)=POS_{g}(Y)\cap BND_{g}(Y)=NEG_{g}(Y)\cap BND_{g}(Y)=\emptyset$. That is, we divide the universe of issues into three disjoint blocks with respect to $Y$.

\begin{theorem}
Let $U$ be the universe of agents, $V$ the universe of issues, and $Y\subseteq U$ a strategy. Then we have
\begin{eqnarray*}
POS_{g}(Y)&=&\{c\in V\mid g^{+}(c)\subseteq Y\}\cap\{c\in V\mid g^{-}(c)\cap Y^{C}\neq \emptyset\};\\
NEG_{g}(Y)&=&\{c\in V\mid g^{-}(c)\subseteq Y\}\cap\{c\in V\mid g^{+}(c)\cap Y^{C}\neq \emptyset\};\\
BND_{g}(Y)&=&POS^{C}_{g}(Y)\cap NEG^{C}_{g}(Y).
\end{eqnarray*}
\end{theorem}

{\bf{Proof:}} Since $(\{c\in V\mid g^{-}(c)\subseteq Y\})^{C}=\{c\in V\mid g^{-}(c)\cap Y^{C}\neq \emptyset\}$,
by Definition 4.3, we have $POS_{g}(Y)=\{c\in V\mid g^{+}(c)\subseteq Y\}-\{c\in V\mid g^{-}(c)\subseteq Y\}=\{c\in V\mid g^{+}(c)\subseteq Y\}\cap (\{c\in V\mid g^{-}(c)\subseteq Y\})^{C}=\{c\in V\mid g^{+}(c)\subseteq Y\}\cap\{c\in V\mid g^{-}(c)\cap Y^{C}\neq \emptyset\}.$ Furthermore, since $(\{c\in V\mid g^{+}(c)\subseteq Y\})^{C}=\{c\in V\mid g^{+}(c)\cap Y^{C}\neq \emptyset\}$,
by Definition 4.3, we have $NEG_{g}(Y)=\{c\in V\mid g^{-}(c)\subseteq Y\}-\{x\in V\mid g^{+}(c)\subseteq Y\}=\{c\in U\mid g^{-}(c)\subseteq Y\}\cap (\{c\in V\mid g^{+}(c)\subseteq Y\})^{C}=\{c\in V\mid g^{-}(c)\subseteq Y\}\cap\{c\in V\mid g^{+}(c)\cap Y^{C}\neq \emptyset\}.$ Finally, we have $BND_{g}(Y)=V-POS_{g}(Y)\cup NEG_{g}(Y)=(POS_{g}(Y)\cup NEG_{g}(Y))^{C}=POS^{C}_{g}(Y)\cap NEG^{C}_{g}(Y).$ $\hspace{1cm}\Box$

Theorem 4.4 illustrates that the agreement strategy $POS_{g}(Y)$ is equal to the intersection of $\{c\in V\mid g^{+}(c)\subseteq Y\}$ and $\{c\in V\mid g^{-}(c)\cap Y^{C}\neq \emptyset\}$, the disagreement strategy $NEG_{g}(Y)$ is equal to the intersection of $\{c\in V\mid g^{-}(c)\subseteq Y\}$ and $\{c\in V\mid g^{+}(c)\cap Y^{C}\neq \emptyset\}$, and the neutral strategy $BND_{g}(Y)$ is equal to the intersection of $POS^{C}_{g}(Y)$ and $NEG^{C}_{g}(Y)$.
Furthermore, if $c_{i}(x)= +1$ and $c_{j}(x)=-1$ for $c_{i}\in POS_{g}(Y)$ and $c_{j}\in NEG_{g}(Y)$, then $x\in Y^{C}$; if $c_{i}(x)=-1$ for $c\in POS_{g}(Y)$ and some $x\in Y$, then there exists $y\in Y^{C}$ such that $c_{i}(x)=-1$;
if $c_{j}(x)=+1$ for $c_{j}\in NEG_{g}(Y)$ and some $x\in Y$, then there exists $y\in Y^{C}$ such that $c_{j}(x)=+1$.

\begin{theorem}
Let $U$ be the universe of agents, $V$ the universe of issues, and $Y\subseteq U$ an agent group. Then we have
\begin{eqnarray*}
&&POS_{g}(Y)=\{c\in V\mid g^{+}(c)\subseteq Y\}\cap\{c\in V\mid \neg(g^{-}(c)\subseteq Y)\};\\
&&NEG_{g}(Y)=\{c\in V\mid g^{-}(c)\subseteq Y\}\cap\{c\in V\mid \neg(g^{+}(c)\subseteq Y)\};\\
&&BND_{g}(Y)=(\{c\in V\mid g^{+}(c)\subseteq Y\}\cap\{c\in V\mid g^{-}(c)\subseteq Y\})\cup (\{c\in V\mid \neg(g^{+}(c)\subseteq Y)\}\cap\{c\in V\mid \neg(g^{-}(c)\subseteq Y)\}).
\end{eqnarray*}
\end{theorem}

{\bf{Proof:}} It is straightforward by Theorem 4.4.$ \hspace{8.5cm}\Box$

Theorem 4.5 illustrates that the agreement, disagreement and neutral strategies of the agent group $Y$ are defined uniformly by using the set inclusion. Concretely, for $Y\subseteq U$ and $c\in V$, if $g^{+}(c)\subseteq Y$ and $\neg(g^{-}(c)\subseteq Y)$, then $c\in POS_{g}(Y)$; if $\neg(g^{+}(c)\subseteq Y)$ and $g^{-}(c)\subseteq Y$, then $c\in NEG_{g}(Y)$; if $g^{+}(c)\subseteq Y$ and $g^{-}(c)\subseteq Y$, then $c\in BND_{g}(Y)$; if $\neg(g^{+}(c)\subseteq Y)$ and $\neg(g^{-}(c)\subseteq Y)$, then $c\in BND_{g}(Y)$.

Suppose $U$ is the universe of agents, $V$ is the universe of issues, $L_{a}=L_{r}=\{F,T\}$ with $F\preceq T$, $L^{+}_{a}=L^{-}_{r}=\{T\}$, $c\in V$, and $Y\subseteq U$ is an agent group. Then the acceptance evaluation $\omega_{r}(c,Y)$ and the rejection evaluation $\omega_{r}(c,Y)$ are defined as follows:
$$\omega_{a}(c,Y)=\left\{
\begin{array}{ccc}
T,&{\rm if}& g^{+}(c)\subseteq Y,\\
F,&{\rm if}& \neg(g^{+}(c)\subseteq Y),\\
\end{array}
\right. \hspace{0.1cm} \omega_{r}(c,Y)=\left\{
\begin{array}{ccc}
T,&{\rm if}& g^{-}(c)\subseteq Y^{C},\\
F,&{\rm if}& \neg(g^{-}(c)\subseteq Y^{C}).\\
\end{array}
\right.$$
Furthermore, by Theorem 4.5, we have the results by using $\omega_{r}(c,Y)$ and $\omega_{r}(c,Y)$ as follows:
\begin{eqnarray*}
POS_{g}(X)=POS_{(\omega_{a},\omega_{r})}(Y)
&=&\{c\in V\mid \omega_{a}(c,Y)\in L^{+}_{a}\wedge \omega_{r}(c,Y)\notin L^{-}_{r}\};\\
NEG_{g}(Y)=NEG_{(\omega_{a},\omega_{r})}(Y)
&=&\{c\in V\mid \omega_{a}(c,Y)\notin L^{+}_{a}\wedge \omega_{r}(c,Y)\in L^{-}_{r}\};\\
BND_{g}(Y)=BND_{(\omega_{a},\omega_{r})}(Y)
&=&(POS_{(\omega_{a},\omega_{r})}(Y)\cup NEG_{(\omega_{a},\omega_{r})}(Y))^{C}\\
&=&\{c\in V\mid (\omega_{a}(c,Y)\notin L^{+}_{a}\wedge \omega_{r}(c,Y)\notin L^{-}_{r})\vee (\omega_{a}(c,Y)\in L^{+}_{a}\wedge \omega_{r}(c,Y)\in L^{-}_{r})\}.
\end{eqnarray*}

Belows, we employ the acceptance model $(A^{\ast},\overline{A^{\ast}})$ and the rejection model $(R^{\ast},\overline{R^{\ast}})$ to interpret $POS_{g}(X)$, $NEG_{g}(X)$ and $BND_{g}(X)$.
Firstly, we provide the acceptance region $POS_{g^{+}}(Y)$ and the non-acceptance region $NPOS_{g^{+}}(Y)$ of the agent group $Y\subseteq U$ in  $(A^{\ast},\overline{A^{\ast}})$ as follows:
\begin{eqnarray*}
POS_{g^{+}}(Y)
=\{c\in V\mid g^{+}(c)\subseteq Y\} \hspace{0.1cm}\text{ and }\hspace{0.1cm}
NPOS_{g^{+}}(Y)
=\{c\in V\mid g^{+}(c)\nsubseteq Y\}.
\end{eqnarray*}
For an issue $c\in V$, we have two-way decisions with respect to an agent group $Y$ as follows:
$(A^{\ast})$ if $g^{+}(c)\subseteq Y$, then we take an acceptance action, i.e. $c\in POS_{g^{+}}(Y)$;
$(\overline{A^{\ast}})$ if $g^{+}(c)\nsubseteq Y$, then we take a non-acceptance action, i.e. $c\in NPOS_{g^{+}}(Y)$. The acceptance rule $(A^{\ast})$ puts issues into an acceptance region, and the non-acceptance rule $(\overline{A^{\ast}})$ classifies issues into a non-acceptance region. Secondly, we provide the rejection region $NEG_{g^{-}}(Y)$ and the non-rejection region $NNEG_{g^{-}}(Y)$ of the agent group $Y\subseteq U$ in $(R^{\ast},\overline{R^{\ast}})$ as follows:
\begin{eqnarray*}
NEG_{g^{-}}(Y)
=\{c\in V\mid g^{-}(c)\subseteq Y\} \hspace{0.1cm}\text{ and }\hspace{0.1cm}
NNEG_{g^{-}}(X)
=\{c\in V\mid g^{-}(c)\nsubseteq Y\}.
\end{eqnarray*}
For an issue $c\in V$, we make two-way decisions with respect to an agent group $Y$ as follows:
$(R^{\ast})$ if $g^{-}(c)\subseteq Y$, then we take a rejection action, i.e. $c\in NEG_{g^{-}}(Y)$;
$(\overline{R^{\ast}})$ if $g^{-}(c)\nsubseteq Y$, then we take a non-rejection action, i.e. $c\in NNEG_{g^{-}}(y)$. The rejection rule $(R^{\ast})$ classifies issues into a rejection region, and the non-rejection rule $(\overline{R^{\ast}})$ puts issues into a non-rejection region. Thirdly, we have the three-way decision rules for the issue $c\in V$ by combining $(A^{\ast},\overline{A^{\ast}})$ and $(R^{\ast},\overline{R^{\ast}})$ as follows:
$(P)$ if $g^{+}(c)\subseteq Y$ and $g^{-}(c)\nsubseteq Y$, then $c\in POS_{g}(Y)$;
$(R)$ if $g^{-}(c)\subseteq Y$ and $g^{+}(c)\nsubseteq Y$, then $c\in NEG_{g}(Y)$;
$(B)$ if $(g^{+}(c)\subseteq Y\wedge g^{-}(c)\subseteq Y)$ or $(g^{+}(c)\nsubseteq Y\wedge g^{-}(c)\nsubseteq Y)$, then $c\in BND_{g}(Y)$.

We see that $POS_{g}(Y)=POS_{g^{+}}(Y)\cap NNEG_{g^{-}}(Y)$, $NEG_{g}(Y)=NPOS_{g^{+}}(Y)\cap NEG_{g^{-}}(Y)$ and $BND_{g}(Y)=(POS_{g^{+}}(Y)\cap NEG_{g^{-}}(Y))\cup (NPOS_{g^{+}}(Y)\cap NNEG_{g^{-}}(Y))$. Meanwhile, we depict the three-way decision rules for the issue $c\in V$ by Table 6 and illustrate that SMZCAs are special cases of TWDCAMs given by Definition 4.2.

\begin{table}[H]\renewcommand{\arraystretch}{1.5}
\caption{Interpretations of $POS_{g}(Y),NEG_{g}(Y)$ and $BND_{g}(Y)$.}
\tabcolsep0.38in
\begin{tabular}{ |c |c |c|}
\hline
 \diagbox{$(A^{\ast},\overline{A^{\ast}})$}{$(R^{\ast},\overline{R^{\ast}})$}&$\omega_{a}(c,Y)=T\in L^{-}_{r}$ &$ \omega_{a}(c,Y)=F\notin L^{-}_{r}$ \\\hline
\multirow{2}*{$\omega_{a}(c,Y)=T\in L^{+}_{a}$}& $c\in BND_{g}(Y)$ &$c\in POS_{g}(Y)$ \\
& (non-commitment) & (acceptance) \\\hline
\multirow{2}*{$\omega_{a}(c,Y)=F\notin L^{+}_{a}$}&$c\in NEG_{g}(Y)$ &$c\in BND_{g}(Y)$ \\
& (rejection) & (non-commitment)\\\hline
\end{tabular}
\end{table}

\subsection{FQWCAMs for Trisecting the Universe of Issues}

In this section, we investigate Fan, Qi and Wei's conflict analysis model\cite{Fan} for dividing issue sets into three disjoint blocks.

Suppose $U$ is the universe of agents, $V$ is the universe of issues, $([0,1],\leq)$ is a totally ordered set, $Y\subseteq U$ is an agent group, $D(g^{+}(c)\mid Y)=\frac{|g^{+}(c)\cap Y|}{|Y|}$ and $
D(g^{-}(c)\mid Y)=\frac{|g^{-}(c)\cap Y|}{|Y|}$. Then the functions $\psi_{a}$ and $\psi_{r}$ are defined as follows:
\begin{eqnarray*}
&&\psi_{a}: U\times P(V)\longrightarrow [0,1],\hspace{0.25cm}\psi_{a}(c,Y)=D(g^{+}(c)\mid Y)=\frac{|g^{+}(c)\cap Y|}{|Y|};\\
&&\psi_{r}: U\times P(V)\longrightarrow [0,1],\hspace{0.25cm}\psi_{r}(c,Y)=D(g^{-}(c)\mid Y)=\frac{|g^{-}(c)\cap Y|}{|Y|}.
\end{eqnarray*}

The functions $\psi_{a}$ and $\psi_{r}$ are mappings from $V\times P(U)$ to $[0,1]$, and $\psi_{a}(c,Y)$ evaluates the extent to which the agent group $Y$ accepts the issue $c$ and $\psi_{r}(c,Y)$ evaluates the extent to which the agent group $Y$ rejects the issue $c$. Furthermore, if we take $L_{a}=L_{r}=[0,1]$ and $\preceq_{a}=\preceq_{r}=\leq$, then the functions $\psi_{a}$ and $\psi_{r}$ are special acceptance evaluation and rejection evaluation given by Definition 4.1, respectively.

\begin{definition}\cite{Fan}
Let $U$ be the universe of agents, $V$ the universe of issues, $(\alpha,1]$ the designated values for acceptance, $(\beta,1]$ the designated values for rejection, and $Y\subseteq U$ an agent group. Then the $(\alpha,\beta)-$agreement, $(\alpha,\beta)-$disagreement and $(\alpha,\beta)-$neutral strategies $POS_{(\alpha,\beta)}(Y)$, $NEG_{(\alpha,\beta)}(Y)$ and $BND_{(\alpha,\beta)}(Y)$ of $Y$ are defined as follows:
\begin{eqnarray*}
&&POS_{(\psi_{a},\psi_{r})}(Y)=\{c\in V\mid \psi_{a}(c,Y)\in (\alpha,1]\wedge \psi_{r}(c,Y)\notin (\beta,1]\};\\
&&NEG_{(\psi_{a},\psi_{r})}(Y)=\{c\in V\mid \psi_{a}(c,Y)\notin (\alpha,1]\wedge \psi_{r}(c,Y)\in (\beta,1]\};\\
&&BND_{(\psi_{a},\psi_{r})}(Y)=V-POS_{(\psi_{a},\psi_{r})}(Y)\cup NEG_{(\psi_{a},\psi_{r})}(Y).
\end{eqnarray*}
\end{definition}

We employ two functions $\psi_{a}(c,Y)$ and $\psi_{r}(c,Y)$ to trisect the universe of issues into $POS_{(\psi_{a},\psi_{r})}(Y)$, $NEG_{(\psi_{a},\psi_{r})}(Y)$ and $BND_{(\psi_{a},\psi_{r})}(Y)$ with respect to $Y$. Especially, we have $POS_{(\psi_{a},\psi_{r})}(Y)\cup NEG_{(\psi_{a},\psi_{r})}(Y)\cup BND_{(\psi_{a},\psi_{r})}(Y)=U$ and $POS_{(\psi_{a},\psi_{r})}(Y)\cap NEG_{(\psi_{a},\psi_{r})}(Y)=POS_{(\psi_{a},\psi_{r})}(Y)\cap BND_{(\psi_{a},\psi_{r})}(Y)=NEG_{(\psi_{a},\psi_{r})}(Y)\cap BND_{(\psi_{a},\psi_{r})}(Y)=\emptyset$.
Furthermore,
if $L^{+}_{a}=(\alpha,1]$, $L^{+}_{r}=(\beta,1]$, $\psi_{a}(c,Y)=\frac{|g^{+}(c)\cap Y|}{|Y|}$, and $\psi_{r}(c,Y)=\frac{|g^{-}(c)\cap Y|}{|Y|}$, then we have
\begin{eqnarray*}
&&POS_{(\psi_{a},\psi_{r}))}(Y)=POS_{(\omega_{a},\omega_{r})}(Y)=\{c\in V\mid \omega_{a}(c,Y)\in L^{+}_{a}\wedge \omega_{r}(c,Y)\notin L^{-}_{a}\};\\
&&NEG_{(\psi_{a},\psi_{r})}(Y)=NEG_{(\omega_{a},\omega_{r})}(Y)=\{c\in V\mid \omega_{a}(c,Y)\notin L^{+}_{a}\wedge \omega_{r}(c,Y)\in L^{-}_{a}\};\\
&&BND_{(\psi_{a},\psi_{r})}(Y)=BND_{(\omega_{a},\omega_{r})}(Y)
=\{c\in V\mid (\omega_{a}(c,Y)\notin L^{+}_{a}\wedge \omega_{r}(c,Y)\notin L^{-}_{r})\vee (\omega_{a}(c,Y)\in L^{+}_{a}\wedge \omega_{r}(c,Y)\in L^{-}_{r})\}.
\end{eqnarray*}

Belows, we take the acceptance model $(A^{\ast},\overline{A^{\ast}})$ and the rejection model $(R^{\ast},\overline{R^{\ast}})$ to interpret $POS_{(\psi_{a},\psi_{r})}(Y)$, $NEG_{(\psi_{a},\psi_{r})}(Y)$ and $BND_{(\psi_{a},\psi_{r})}(Y)$.
Firstly, we provide the acceptance region $POS_{\psi_{a}}(Y)$ and the non-acceptance region $NPOS_{\psi_{a}}(Y)$ of the agent group $Y\subseteq U$ in $(A^{\ast},\overline{A^{\ast}})$ as follows:
\begin{eqnarray*}
POS_{\psi_{a}}(Y)
=\{c\in V\mid \psi_{a}(c,Y)\in (\alpha,1]\} \hspace{0.1cm}\text{ and }\hspace{0.1cm}
NPOS_{\psi_{a}}(X)
=\{c\in V\mid \psi_{a}(c,Y)\notin (\alpha,1]\}.
\end{eqnarray*}
For an issue $c\in V$, we make two-way decisions with respect to an agent group $Y$ as follows:
$(A^{\ast})$ if $\psi_{a}(c,Y)\in (\alpha,1]$, then take an acceptance action, i.e. $c\in POS_{\psi_{a}}(Y)$;
$(\overline{A^{\ast}})$ if $\psi_{a}(c,Y)\notin (\alpha,1]$, then take a non-acceptance action, i.e. $c\in NPOS_{\psi_{a}}(Y)$. The acceptance rule $(A^{\ast})$ classifies issues into an acceptance region, and the non-acceptance rule $(\overline{A^{\ast}})$ puts issues into a non-acceptance region.
Secondly, we give the rejection region $NEG_{\psi_{a}}(Y)$ and the non-rejection region $NNEG_{\psi_{a}}(Y)$ of the agent group $Y\subseteq U$ in $(R^{\ast},\overline{R^{\ast}})$ as follows:
\begin{eqnarray*}
NEG_{\psi_{r}}(Y)
=\{c\in V\mid \psi_{r}(x,Y)\in (\beta,1]\} \hspace{0.1cm}\text{ and }\hspace{0.1cm}
NNEG_{\psi_{r}}(Y)
=\{c\in V\mid \psi_{r}(x,Y)\notin (\beta,1]\}.
\end{eqnarray*}
For an issue $c\in V$, we make two-way decisions with respect to an agent group $Y$ as follows:
$(A)$ if $\psi_{r}(c,Y)\in (\beta,1]$, then we take a rejection action, i.e. $c\in NEG_{\psi_{r}}(Y)$;
$(\overline{A})$ if $\psi_{r}(c,Y)\notin (\beta,1]$, then we take a non-rejection action, i.e. $c\in NNEG_{\psi_{r}}(Y)$. The rejection rule $(R^{\ast})$ puts issues into an rejection region, and the non-rejection rule $(\overline{R^{\ast}})$ classifies issues into a non-rejection region. Thirdly, we have the three-way decision rules for the issue $c\in V$ by combining $(A^{\ast},\overline{A^{\ast}})$ and $(R^{\ast},\overline{R^{\ast}})$ as follows:
$(P)$ if $\psi_{a}(c,Y)\in (\alpha,1]$ and $\psi_{r}(c,Y)\notin (\beta,1]$, then $x\in POS_{(\psi_{a},\psi_{r})}(Y)$;
$(R)$ if $\psi_{r}(c,Y)\in (\beta,1]$ and $\psi_{a}(c,Y)\notin (\alpha,1]$, then $c\in NEG_{(\psi_{a},\psi_{r})}(Y)$;
$(B)$ if $(\psi_{a}(c,Y)\in (\alpha,1]\wedge\psi_{r}(c,Y)\in (\beta,1])$ or $(\psi_{a}(c,Y)\notin (\alpha,1]\wedge\psi_{r}(c,Y)\notin (\beta,1])$, then $c\in BND_{(\psi_{a},\psi_{r})}(Y)$.

\begin{table}[H]\renewcommand{\arraystretch}{1.5}
\caption{Interpretations of $POS_{(\psi_{a},\psi_{r})}(Y),NEG_{(\psi_{a},\psi_{r})}(Y)$ and $BND_{(\psi_{a},\psi_{r})}(Y)$.}
\tabcolsep0.35in
\begin{tabular}{ |c |c |c|}
\hline
 \diagbox{$(A^{\ast},\overline{A^{\ast}})$}{$(R^{\ast},\overline{R^{\ast}})$}&$\psi_{r}(c,Y)\in L^{-}_{r}=(\beta,1]$ &$\psi_{r}(c,Y)\notin L^{-}_{r}=(\beta,1]$ \\\hline
\multirow{2}*{$\psi_{a}(c,Y)\in L^{+}_{a}=(\alpha,1]$}& $c\in BND_{(\psi_{a},\psi_{r})}(Y)$ &$c\in POS_{(\psi_{a},\psi_{r})}(Y)$\\
& (non-commitment) & (acceptance)  \\\hline
\multirow{2}*{$\psi_{a}(c,Y)\notin L^{+}_{a}=(\alpha,1]$}& $c\in NEG_{(\psi_{a},\psi_{r})}(Y)$ &$c\in BND_{(\psi_{a},\psi_{r})}(Y)$\\
& (rejection) & (non-commitment)\\\hline
\end{tabular}
\end{table}

We observe that $POS_{(\psi_{a},\psi_{r})}(Y)=POS_{\psi_{a}}(Y)\cap NNEG_{\psi_{r}}(Y)$, $NEG_{(\psi_{a},\psi_{r})}(Y)=NPOS_{\psi_{a}}(Y)\cap NEG_{\psi_{r}}(Y)$ and $BND_{(\psi_{a},\psi_{r})}(Y)=(POS_{\psi_{a}}(Y)\cap NEG_{\psi_{r}}(Y))\cup (NPOS_{\psi_{a}}(Y)\cap NNEG_{\psi_{r}}(Y))$. Meanwhile, we depict the three-way decision rules for the issue $c\in V$ by Table 7 and illustrate that FQWCAs are special cases of TWDCAMs given by Definition 4.2.

\section{The relationship among TWDCAMs, FQWCAMs and SMZCAMs}

In this section, we investigate the relationship among TWDCAMs, FQWCAMs and SMZCAMs.

On one hand, there are three types of conflict analysis models TWDCAMs, FQWCAMs and SMZCAMs for trisecting the universes of agents and issues, and six conflict analysis models are given by Definitions 3.2, 3.3, 3.8, 4.2, 4.3 and 4.6 in Sections 3 and 4. Concretely, we employ Definitions 3.2, 3.3 and 3.8 to divide the universe of agents into three disjoint blocks; we apply Definitions 4.2, 4.3 and 4.6 to trisect the universe of issues. Especially, Definitions 3.3 and 3.8 are special cases of Definition 3.2; Definitions 4.3 and 4.6 are special cases of Definition 4.2.
Therefore, we classify the six conflict analysis models into two categories as follows: $\{\text{Definition 3.2, Definition 3.3, Definition 3.8}\}$ and $\{\text{Definition 4.2, Definition 4.3, Definition 4.6}\}$, which are depicted by Table 8, where $\surd$ denotes that the conflict analysis model satisfies the property in the column.

\begin{table}[H]\renewcommand{\arraystretch}{1.5}
\caption{Classification of Six Conflict Analysis Models}
\tabcolsep0.145in
\begin{tabular}{ | c | c | c | c | c |}
\hline
 &\text{Divide Agents}  & \text{Divide Issues} &\text{Qualitative Models}  &\text{Quantitative Models} \\\hline
\text{Definition 3.2} & $\surd$ & & $\surd$ &\\\hline
\text{Definition 3.3} & $\surd$ & & & $\surd$\\\hline
\text{Definition 3.8} & $\surd$& & $\surd$ &\\\hline
\text{Definition 4.2} & & $\surd$ & $\surd$ &\\\hline
\text{Definition 4.3} & & $\surd$ & & $\surd$\\\hline
\text{Definition 4.6} & & $\surd$ &  $\surd$&\\\hline
\end{tabular}
\end{table}

On the other hand, the conflict analysis models given by Definitions 3.2 and 4.2 employ two evaluation functions to construct the agreement, disagreement and neutral subsets of a strategy and the agreement, disagreement and neutral strategies of an agent group, respectively;
the conflict analysis models given by 3.3 and 4.3 apply the set inclusion to define the agreement, disagreement and neutral subsets of a strategy and the agreement, disagreement and neutral strategies of an agent group, respectively; the conflict analysis models given by Definitions 3.8 and 4.6 take the including degree function to construct the agreement, disagreement and neutral subsets of a strategy and the agreement, disagreement and neutral strategies of an agent group, respectively. Furthermore, the set inclusion and including degree function are considered as special cases of the evaluation functions, and Definitions 3.2 and 4.2 are generalizations of Definitions 3.3 and 3.8 and Definitions 4.3 and 4.6, respectively. Especially, the conflict analysis models given by Definitions 3.3 and 4.3 are qualitative models, and the conflict analysis models given by Definitions 3.8 and 4.6 are quantitative models, which are depicted by Table 8, where $\surd$ denotes that the conflict analysis model satisfies the property in the column.

\section{Conclusions}

Three-way decision theory is a powerful mathematical tool for handling uncertainty in conflict analysis decision making problems. In this paper, we have provided a type of TWDCAMs for trisecting the universe of agents and employed a pair of two-way decisions models to interpret the three-way decisions rules of an agent. Moreover, we have established another type of TWDCAMs for trisecting the universe of issues and interpreted the three-way decisions rules of an issue with a couple of two-way decisions models. Finally, we have interpreted FQWCAMs and SMZCAMs with a pair of two-day decisions models and illustrated that FQWCAMs and SMZCAMs are special cases of TWDCAMs.

In the future, we will study how to provide the designated values for acceptance and designed values for rejection, and provide effective algorithms for trisecting the universes of agents and issues. Furthermore, there are a lot of dynamic situation tables in practical situations, and we will investigate how to trisect the universes of agents and issues in dynamic situation tables.

\section*{Acknowledgments}

We would like to thank the anonymous reviewers very much for their
professional comments and valuable suggestions. This work is
supported by the National Natural Science Foundation of China (Nos. 61603063,61673301,11771059,61573255),
Hunan Provincial Natural Science Foundation of China(Nos.2018JJ3518, 2018JJ2027), the Scientific Research Fund of Hunan Provincial Key Laboratory of Mathematical Modeling and Analysis
in Engineering (No. 2018MMAEZD10).

\end{document}